\documentclass[sigconf]{acmart}

\AtBeginDocument{%
  }

\copyrightyear{2026}
\acmYear{2026}
\setcopyright{cc}
\setcctype{by}
\acmConference[RichMediaGAI '26]{The fourth International Workshop on Rich Media with Generative AI }{November 10--14, 2026}{Rio de Janeiro, Brazil}
\acmBooktitle{The fourth International Workshop on Rich Media with Generative AI (RichMediaGAI '26), November 10--14, 2026, Rio de Janeiro, Brazil}
\acmDOI{10.1145/3841458.3841538}
\acmISBN{979-8-4007-2947-8/2026/11}

\begin{document}

\title{The RenAIssance: Caption-Guided Diffusion Models for Historical-Art Visual Media}

\author{Zhanyu Tuo}
\affiliation{
  \institution{Sorbonne University}
  \city{Paris}
  \country{France}}
\email{zhanyu.tuo@sorbonne-universite.fr}

\begin{abstract}
Generative AI can expand cultural collections into prompt-responsive visual media, but historical artworks rarely provide the paired descriptions needed to adapt modern image generators. We present \textit{The RenAIssance}, a scalable pipeline that turns a collection of historical European artworks into a reusable text-to-image media resource. The pipeline generates scene-level captions with BLIP and uses the resulting approximately 15,000 image--text pairs from DEArt to train a compact LoRA adapter for Stable Diffusion v1.5. The adapter shifts generated images toward the historical-art distribution while retaining the base model's ability to render novel concepts. On historical and contemporary prompts of different lengths, it improves FID to the DEArt reference distribution in every setting with only a modest decrease in CLIP text--image similarity. Qualitative results show that the same adapter can render both period-appropriate subjects and contemporary concepts in a coherent historical visual language. By decoupling a reusable style representation from the base generator, the approach offers a practical building block for controllable historical-art media creation, interactive authoring, and rich-media experiences.
\end{abstract}



\maketitle

\section{Introduction}
Visual-media systems increasingly use generative models to create, adapt, and personalize visual assets. This capability can make historical artworks accessible in interactive experiences, educational storytelling, virtual exhibitions, creative production, and so on. Yet the images in museum-scale collections are typically accompanied by sparse catalog metadata instead of the detailed descriptions required by prompt-conditioned generators. A usable generative-media pipeline must therefore connect collection representation, semantic conditioning, efficient adaptation, and controllable rendering.

Among recent advances in generative AI, diffusion models have emerged as the dominant paradigm for image synthesis. Starting from the Denoising Diffusion Probabilistic Model (DDPM) framework \cite{ho2020ddpm} and its subsequent improvements \cite{nichol2021improved}, diffusion-based methods have demonstrated remarkable capabilities in generating realistic and controllable visual content. Among these approaches, diffusion models \cite{saharia2022photorealistic,podell2024sdxl,rombach2022high} have emerged as the dominant paradigm for text-to-image generation. Despite their success, most large-scale diffusion models are trained on web-scale image collections that predominantly reflect contemporary visual culture and internet imagery. Consequently, these models are not explicitly specialized for historical artistic domains and may struggle to consistently reproduce the stylistic characteristics associated with historical artwork collections.

The application of generative AI to historical art remains relatively underexplored compared with contemporary artistic styles. Historical artwork collections provide rich visual representations of artistic traditions, cultural narratives, and aesthetic practices developed over centuries. However, adapting modern text-to-image models to these domains presents several challenges. First, historical artwork datasets are typically much smaller than modern web-scale datasets. Second, many collections lack paired textual descriptions, making them difficult to use for training text-conditioned diffusion models. As a result, generating images that consistently reflect the visual characteristics of historical artworks remains a challenging task.

Datasets such as DEArt \cite{reshetnikov2022deart} provide valuable resources for studying historical artistic styles by collecting and organizing artworks spanning multiple periods in Europe. However, DEArt was originally designed for artwork understanding and object detection tasks and therefore contains object labels and bounding-box annotations rather than detailed image captions. This limits its direct applicability for text-to-image diffusion model adaptation, where paired image--text data are typically required.

To address this limitation, we propose \textit{The RenAIssance}, an effective and scalable strategy for adapting diffusion models to historical artwork collections that lack paired image–text annotations. Specifically, we employ the Bootstrapping Language-Image Pre-training (BLIP) model \cite{li2022blip,li2023blip} to automatically generate descriptive captions for artworks contained in the DEArt dataset. These image--text pairs are subsequently used to adapt Stable Diffusion \cite{rombach2022high} through Low-Rank Adaptation (LoRA) \cite{hu2022lora}, enabling efficient specialization toward historical artistic styles while preserving the general generation capabilities of the pre-trained model.

Unlike prior approaches that primarily focus on contemporary artistic styles or style transfer from a small number of reference images, our framework learns stylistic characteristics from a large collection of historical artworks. We further investigate whether the adapted model can generalize beyond the training distribution by generating both historically relevant subjects and contemporary concepts that are absent from the training dataset. To evaluate the effectiveness of the proposed framework, we design experiments using both short and detailed prompts and consider historical as well as contemporary content descriptions. The generated images are evaluated quantitatively using CLIP similarity \cite{radford2021learning} and Fréchet Inception Distance (FID) \cite{heusel2017gans}, complemented by qualitative visual analysis.

Our contributions are:
\begin{itemize}
\item a scalable caption-to-adapter pipeline that transforms unpaired historical artworks into a prompt-addressable visual-media corpus;
\item a parameter-efficient collection representation that specializes a text-to-image generator without replacing its general semantic knowledge; and
\item an evaluation showing improved alignment with a historical-art reference distribution across historical and contemporary prompts, with qualitative evidence of controllable visual-media creation.
\end{itemize}

\begin{figure}[ht]
\centering
\includegraphics[width=\columnwidth]{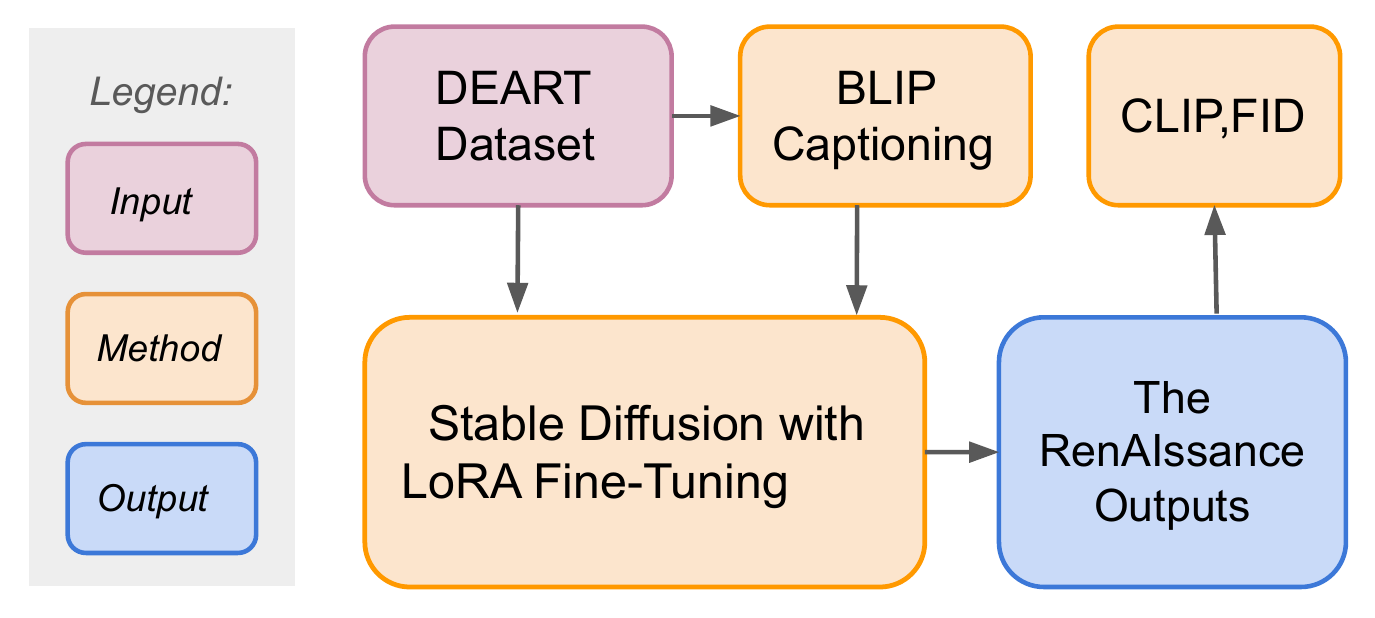}
\caption{Overview of the proposed pipeline. BLIP is first used to automatically generate captions for artworks from the DEArt dataset. The resulting image--text pairs are employed to fine-tune a Stable Diffusion model using LoRA. The adapted model is subsequently evaluated using historical and contemporary prompts of varying lengths, and the generated images are assessed using CLIP similarity, FID, and qualitative visual comparisons.}
\label{fig:pipeline}
\end{figure}

\section{Related Work}

\subsection{Diffusion Models}

Recent advances in large-scale vision--language datasets such as LAION-5B \cite{schuhmann2022laion} have significantly accelerated the development of text-to-image generative models. Earlier generative approaches, including Variational Autoencoders (VAEs) \cite{kingma2013auto} and Generative Adversarial Networks (GANs) \cite{goodfellow2020generative}, demonstrated the feasibility of image synthesis but often suffered from limitations such as blurry reconstructions, training instability, or mode collapse.

Modern diffusion models originate from the Denoising Diffusion Probabilistic Model (DDPM) framework proposed by Ho \textit{et al.} \cite{ho2020ddpm}, which formulates image generation as an iterative denoising process. Nichol and Dhariwal \cite{nichol2021improved} subsequently introduced improved sampling and variance learning strategies that significantly enhanced generation quality and sampling efficiency. 

Building upon these foundations, Stable Diffusion \cite{rombach2022high} introduced latent diffusion models that perform denoising in a compressed latent space, substantially reducing computational requirements while maintaining high image quality. Stable Diffusion XL (SDXL) \cite{podell2024sdxl} further improved image realism, prompt alignment, and compositional generation through larger-scale training and architectural refinements.

Beyond image generation, diffusion models have been successfully applied to a wide range of tasks, including image editing \cite{huang2025photodoodle,zhang2025stablemakeup,zhang2025stable,jia2025d}, video generation and understanding \cite{xi2026omnivdiff,11415357,chen2026transanimate},  and multi-modal content creation. More recently, transformer-based architectures have been introduced to diffusion modeling. Diffusion Transformers (DiT) \cite{peebles2023scalable} replace conventional backbones with transformer architectures, demonstrating improved scalability and generation quality. Subsequent works such as Hunyuan-DiT \cite{li2024hunyuan} and MMFace-DiT \cite{krishnamurthy2026mmface} further explored the integration of diffusion models and transformers for high-fidelity image generation under diverse conditioning signals.
\enlargethispage{10pt}

\subsection{Style-Consistent Image Generation}

The ability to generate images with consistent artistic styles has become an important research direction in generative AI. Early style-transfer approaches focused on transferring visual characteristics from a reference image to a target image, while more recent diffusion-based methods enable style control directly during image generation.

Several tuning-free approaches have been proposed to inject style information into pre-trained diffusion models. IP-Adapter \cite{ye2023ip} introduces image-based conditioning mechanisms that preserve compatibility with text prompts while allowing visual guidance from reference images. StyleAdapter \cite{wang2025styleadapter} extends this idea by learning unified style representations for stylized image generation. Similarly, StyleAlign \cite{wu2021stylealign} investigates style alignment mechanisms to improve style preservation across generated samples. These approaches offer efficient style control without modifying the underlying diffusion model weights; however, they often rely on one or a few reference images and may struggle to capture complex stylistic characteristics spanning an entire artistic movement.

To improve style consistency, parameter-efficient fine-tuning techniques based on LoRA \cite{hu2022lora} have become increasingly popular. By introducing trainable low-rank matrices into pre-trained diffusion models, LoRA enables efficient specialization toward particular artistic styles while requiring only a small fraction of the original model parameters. Recent works such as B-LoRA \cite{frenkel2024implicit} and ConsisLoRA \cite{chen2025consislora} demonstrate that LoRA-based adaptation can significantly improve both content preservation and stylistic consistency compared with tuning-free approaches. As a result, style-specific LoRA modules have become a practical solution for personalized image generation and artistic style adaptation.

\section{Pipeline}

An overview of the proposed \textit{The RenAIssance} is illustrated in Fig. \ref{fig:pipeline}. The objective of the pipeline is to adapt a pre-trained text-to-image diffusion model to generate images that reflect the visual characteristics of historical artworks while preserving the flexibility of prompt-based image synthesis. To achieve this goal, the whole framework consists of three stages: (1) Automatic caption generation, (2) LoRA-based diffusion model adaptation, and (3) Quantitative and qualitative evaluation.

\subsection{Automatic caption generation}

The training data are derived from the Dataset of European Art (DEArt) \cite{reshetnikov2022deart}, which contains more than 15,000 artworks spanning multiple periods and regions of European art history. Although DEArt provides a large collection of curated artworks, it does not include paired textual descriptions suitable for training text-conditioned diffusion models. To address this limitation, we employ BLIP \cite{li2022blip,li2023blip} to automatically generate image captions for each artwork in DEArt.

Specifically, BLIP is used to produce natural-language descriptions that summarize the visual content of each image, including depicted objects, scenes, activities, and artistic attributes. The generated captions are paired with the corresponding artworks to create an image--text dataset compatible with the training of Stable Diffusion models. Compared with manual annotation, this automated approach enables scalable dataset construction while maintaining semantic correspondence between images and textual descriptions.

\subsection{LoRA-based Adaptation}

The resulting image--text pairs are then used to adapt a pre-trained Stable Diffusion model \cite{rombach2022high}. Rather than updating all model parameters, we adopt Low-Rank Adaptation (LoRA) \cite{hu2022lora}, a parameter-efficient fine-tuning strategy that introduces trainable low-rank decomposition matrices into selected layers of the diffusion model. During training, the original model parameters remain frozen while only the additional LoRA parameters are optimized. This substantially reduces memory consumption and computational cost while allowing the model to learn stylistic characteristics from the target artwork collection.

After fine-tuning for 5,000 training iterations, the adapted model is evaluated using prompts designed to assess both style acquisition and prompt generalization. We consider two complementary prompt categories. Historical-content prompts describe subjects commonly found in historical paintings, such as religious scenes, portraits, or historical events, whereas contemporary-content prompts describe modern concepts that are rarely present in historical artworks. This evaluation protocol allows us to investigate whether the learned artistic style can be transferred to novel semantic content beyond the training distribution.

To further analyze prompt sensitivity, each category is evaluated using both short and long detailed textual prompts. Short prompts provide minimal semantic guidance, whereas long prompts contain richer contextual information and stylistic descriptions. Comparing these prompt types enables us to assess the influence of prompt complexity on style consistency and image generation quality.

\subsection{Evaluation Protocol}

The generated images are evaluated using both quantitative and qualitative metrics. We employ FID \cite{heusel2017gans} to measure the distributional similarity between generated images and reference artworks. FID computes the 2-Wasserstein distance between feature distributions extracted from an Inception network, where lower scores indicate greater similarity between the generated and reference image distributions.

In addition, we use CLIP similarity \cite{radford2021learning} to evaluate text--image alignment. Specifically, CLIP measures the semantic consistency between an input prompt and the generated image within a shared vision--language embedding space. Higher CLIP similarity scores indicate stronger correspondence between the textual description and the generated visual content.

Finally, qualitative comparisons are conducted between the original Stable Diffusion model and the LoRA-adapted model. By analyzing generated samples across different prompt categories and prompt lengths, we investigate the extent to which the proposed framework captures stylistic characteristics of historical European artworks while maintaining semantic fidelity to user-provided prompts.

\begin{figure}[ht]
\centering
\includegraphics[width=\columnwidth]{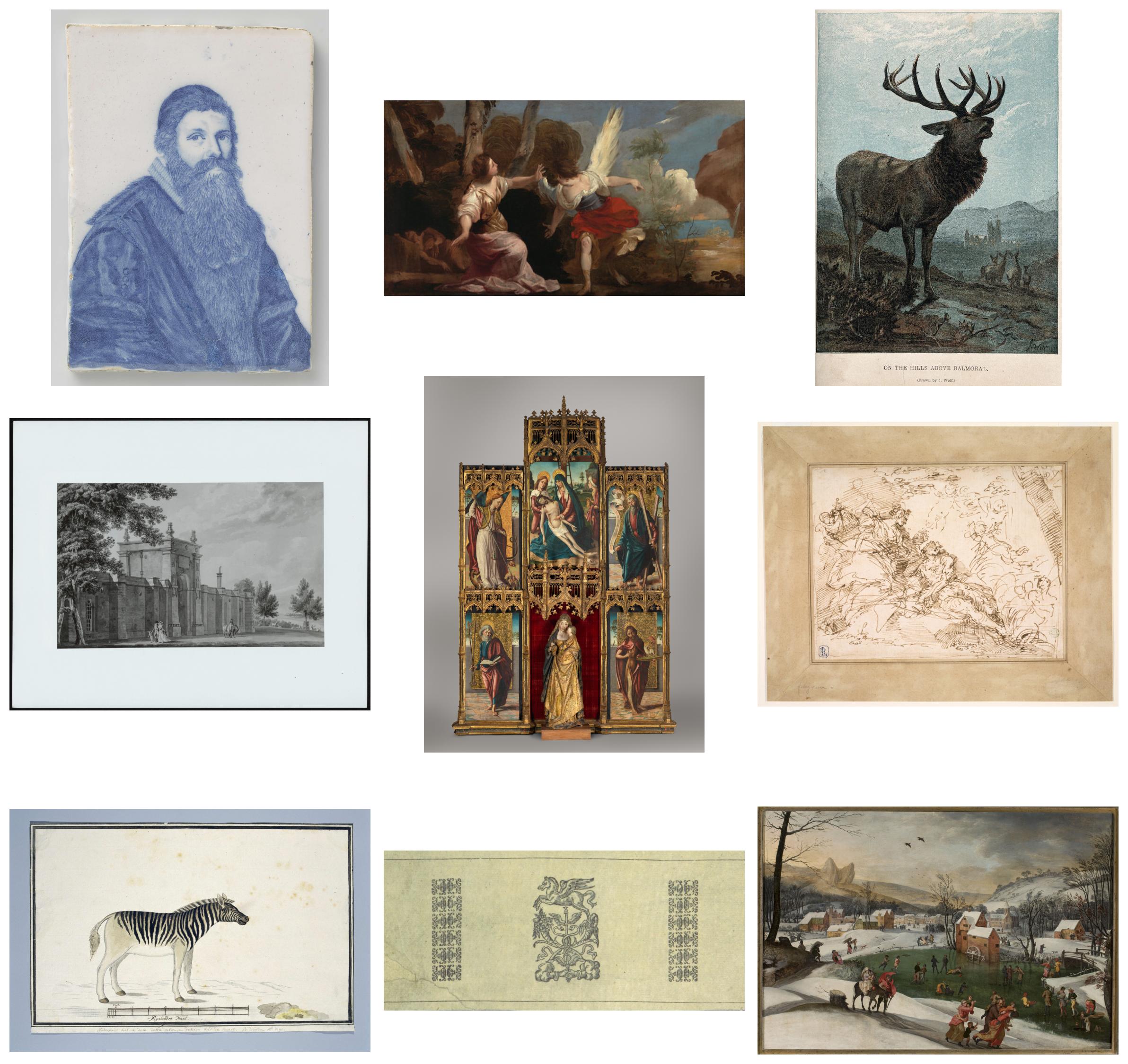}
\caption{Representative samples from the DEArt dataset. The collection contains artworks spanning multiple centuries of European art history and exhibits substantial diversity in subject matter, composition, and artistic style.}
\label{fig:9 samples}
\end{figure}

\section{Dataset Preparation}
\label{sec:dataset}
 
In this work, we employ the Dataset of European Art (DEArt) \cite{reshetnikov2022deart} as the primary source of historical artwork images. DEArt contains more than 15,000 artworks spanning the twelfth to the eighteenth centuries and covers a broad range of artistic periods, subject categories, and visual styles. Originally developed for artwork understanding and object detection, the dataset provides object labels and bounding-box annotations rather than descriptive image captions. Although these annotations are suitable for recognition tasks, they lack the rich semantic descriptions required for training or adapting modern text-conditioned diffusion models.

To overcome this limitation, we construct a paired image--text dataset through automatic caption generation. Specifically, we employ the BLIP vision--language model \cite{li2022blip,li2023blip} to generate natural-language descriptions for every artwork in DEArt. Given an input image, BLIP produces a caption that summarizes the visual content of the artwork, including objects, human activities, scene composition, and contextual information. These automatically generated captions are then paired with their corresponding images to produce a training corpus compatible with Stable Diffusion fine-tuning.

Formally, let
\begin{equation}
\mathcal{I}=\{I_i\}_{i=1}^{N},
\end{equation}
denote the DEArt artwork collection, where \(N\approx15{,}000\). The BLIP captioning model defines a mapping
\begin{equation}
c_i=f_{\mathrm{BLIP}}(I_i),
\end{equation}
where \(c_i\) is the automatically generated natural-language description associated with artwork \(I_i\). The resulting paired image--text dataset is therefore
\begin{equation}
\mathcal{D}
=
\left\{
\left(I_i,c_i\right)
\right\}_{i=1}^{N},
\end{equation}
which serves as the training corpus for subsequent LoRA-based adaptation of the Stable Diffusion model.

\begin{figure}[h]
\centering
\includegraphics[width=\columnwidth]{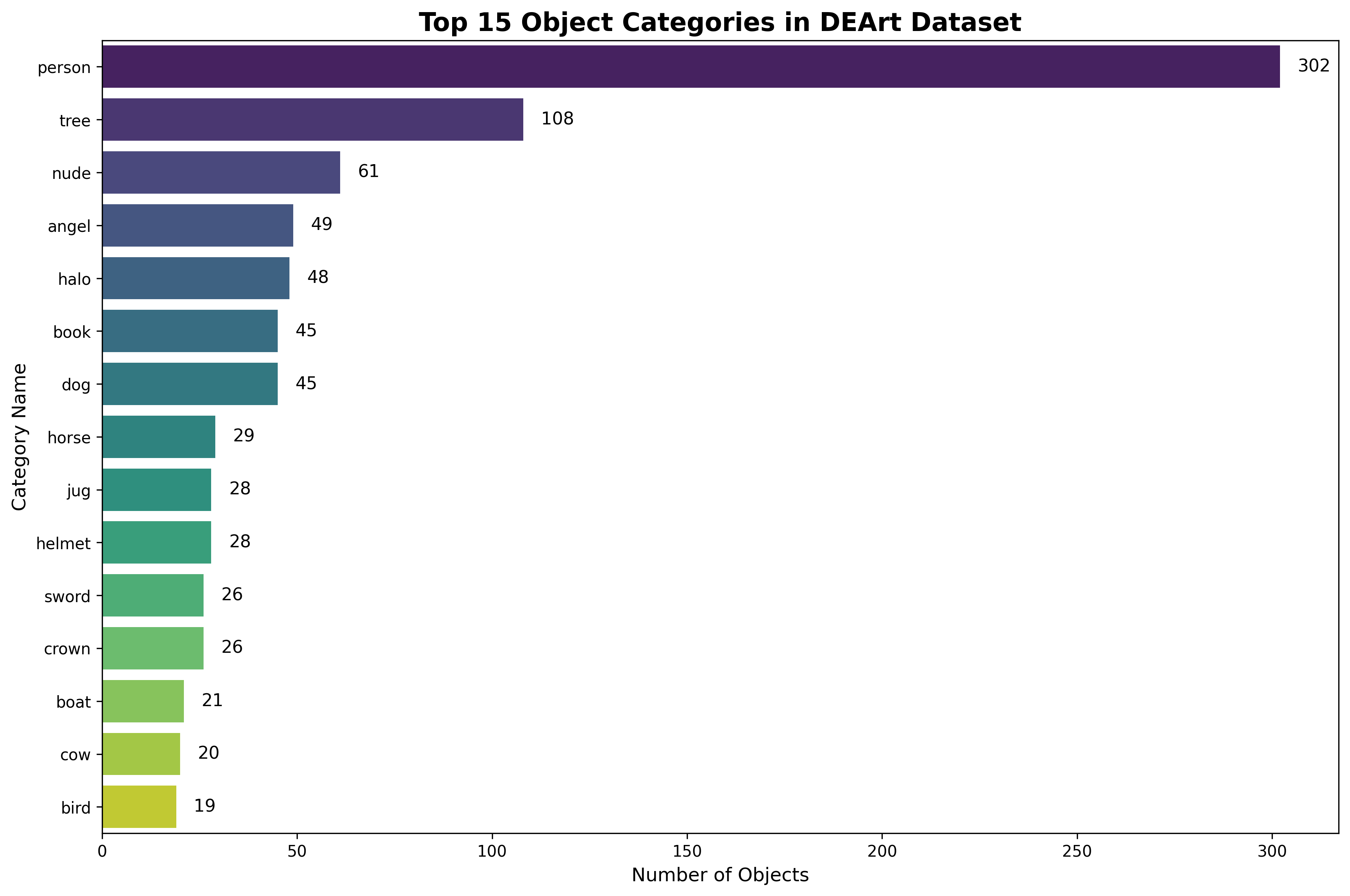}
\caption{Distribution of the fifteen most frequent object categories in the DEArt dataset. The distribution of object categories reflects the semantic coverage of the historical artwork collection.}
\label{fig:distribution}
\end{figure}

Examples of artworks from DEArt are shown in Fig. \ref{fig:9 samples}. The collection exhibits substantial diversity in both semantic content and artistic representation, including portraits, religious scenes, landscapes, architectural structures, and everyday activities. The distribution of the fifteen most frequent object categories is illustrated in Fig. \ref{fig:distribution}, highlighting the semantic coverage of the dataset.

\begin{figure*}[t]
\centering
\includegraphics[width=\textwidth]{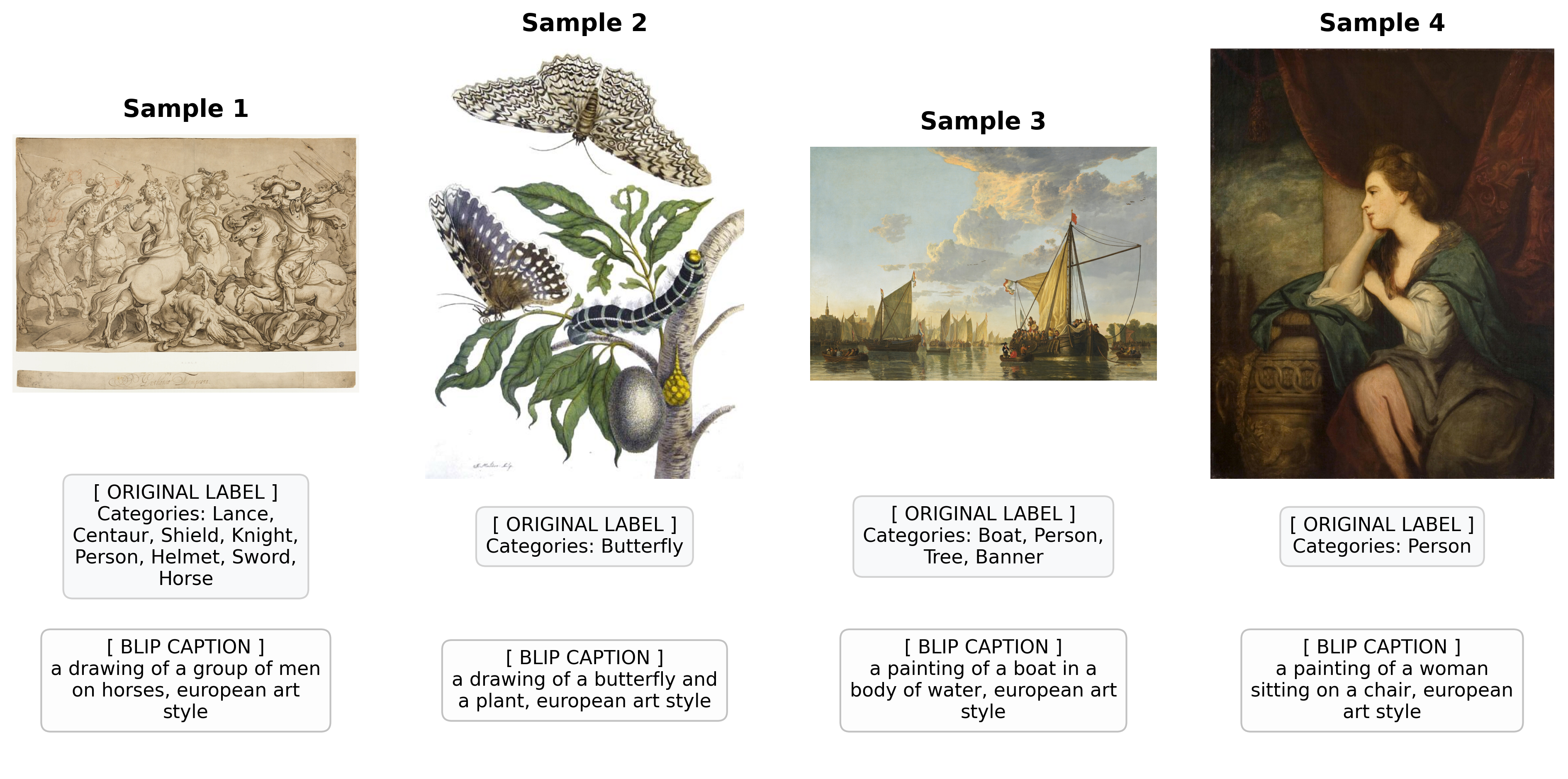}
\caption{Examples of automatic caption generation. The first row shows original artworks from DEArt, the second row presents the original object annotations, and the third row shows captions generated by BLIP. Compared with the original labels, the generated captions provide richer scene-level descriptions, making the dataset suitable for text-to-image diffusion model adaptation.}
\label{fig:caption}
\end{figure*}

Fig. \ref{fig:caption} presents representative examples of the automatically generated captions. Compared with the original object annotations provided by DEArt, the BLIP-generated captions capture substantially richer semantic information and more complete scene-level descriptions. Rather than identifying isolated objects, the generated captions also describe relationships among subjects, actions, and visual attributes.

\section{Experiments}

\subsection{LoRA-Based Adaptation}

We adopt Stable Diffusion v1.5 \cite{rombach2022high} as the base text-to-image generation model and employ LoRA \cite{hu2022lora} for parameter-efficient fine-tuning. Instead of updating all parameters of the diffusion model, LoRA injects trainable low-rank matrices into selected attention layers while keeping the original pre-trained weights frozen. This strategy substantially reduces the number of trainable parameters and computational cost while enabling efficient adaptation to the domain of historical European artworks.

Formally, let \(W_0\in\mathbb{R}^{k\times d}\) denote a pre-trained weight matrix in the Stable Diffusion model. LoRA approximates the weight update \(\Delta W\) using two trainable low-rank matrices \(A\in\mathbb{R}^{r\times d}\) and \(B\in\mathbb{R}^{k\times r}\), where the rank satisfies \(r\ll\min(k,d)\). The adapted weight matrix is therefore expressed as

\begin{equation}
W
=
W_0+\Delta W
=
W_0+BA,
\label{eq:lora}
\end{equation}
where only the low-rank matrices \(A\) and \(B\) are optimized during fine-tuning. By constraining the parameter updates to a low-dimensional subspace, LoRA efficiently learns domain-specific stylistic representations while largely preserving the general image-generation capabilities of the pre-trained diffusion model.

During training on the paired image--text dataset
\(
\mathcal{D}=\{(I_i,c_i)\}_{i=1}^{N}
\)
defined in Sec.~\ref{sec:dataset}, all original Stable Diffusion parameters remain frozen, and only the LoRA parameters
\(
\theta_{\mathrm{LoRA}}
\)
are optimized according to the standard diffusion objective,

\begin{equation}
\theta_{\mathrm{LoRA}}^{*}
=
\arg\min_{\theta_{\mathrm{LoRA}}}
\;
\mathcal{L}_{\mathrm{diffusion}}
\left(
\mathcal{D};
\theta_{\mathrm{LoRA}}
\right),
\label{eq:optimization}
\end{equation}
where \(\mathcal{L}_{\mathrm{diffusion}}\) denotes the denoising objective employed by Stable Diffusion model and \(\theta_{\mathrm{LoRA}}\) represents the trainable LoRA parameters. Here,
\begin{equation}
    \theta_{\rm LoRA}
=
\{A^{(l)},B^{(l)}\}_{l=1}^{L},
\end{equation}
where \(L\) is the number of LoRA-adapted attention layers. Thus, only the low-rank matrices are optimized during training, while the original Stable Diffusion parameters remain fixed. The optimization is performed exclusively on the automatically generated DEArt image--text pairs, allowing the adapted model to acquire stylistic characteristics commonly observed in historical artworks, including painting texture, color palettes, scene composition, and artistic representations, while preserving the semantic controllability inherited from the pre-trained model.

\subsection{Training Configuration}

The LoRA adapters are trained for 5,000 optimization steps using the automatically generated image--text dataset described in Section~\ref{sec:dataset}. To balance computational efficiency and training stability, we employ gradient accumulation during optimization. Specifically, a micro-batch size of four samples is accumulated over four iterations, resulting in an effective batch size of sixteen samples per parameter update.

Optimization is performed using the Adam optimizer \cite{kingma2014adam} with mixed-precision training in the bfloat16 format. Mixed-precision computation reduces GPU memory consumption and accelerates training while maintaining numerical stability. In addition, 8-bit optimizer states are employed to further reduce memory requirements, allowing the complete training procedure to be executed on a single GPU.\enlargethispage{16pt}

\subsection{Implementation Details}
 
Since the DEArt dataset is stored in Parquet format on a shared network file system, data loading was identified as a potential bottleneck during training. To improve training stability, the data-loading pipeline was configured to prioritize reliable sequential access rather than aggressive multiprocessing. This design choice reduced I/O-related interruptions and ensured consistent training throughout the fine-tuning process.

Following adaptation, the resulting model is evaluated using both historical and contemporary prompts with varying levels of textual detail. Quantitative evaluation is performed using CLIP similarity and FID, while qualitative comparisons are used to assess style consistency, visual quality, and prompt alignment.

\section{Results}

\begin{figure*}[h]
\centering
\includegraphics[width=\textwidth]{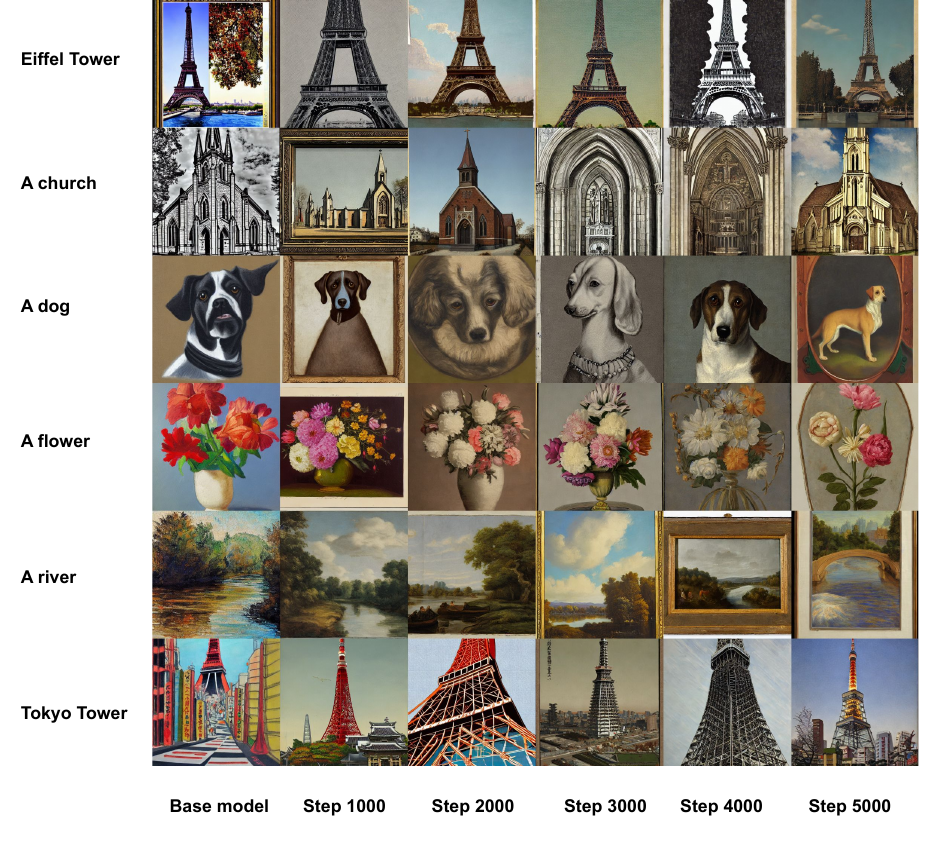}
\caption{Qualitative comparison using short prompts. Columns correspond to the base Stable Diffusion model and LoRA-adapted checkpoints from 1,000 to 5,000 training steps. The adapted model progressively acquires stylistic characteristics associated with historical artworks.}
\label{fig:outputs}
\end{figure*}

The proposed framework aims to adapt a pre-trained diffusion model toward the visual characteristics of historical artworks while preserving its ability to generate diverse content from natural-language prompts. During inference, the learned LoRA parameters are merged with the Stable Diffusion model, enabling style-aware image generation without modifying the original model weights.

To evaluate both stylistic adaptation and prompt generalization, we design experiments using two prompt categories: historical-content prompts and contemporary-content prompts. Historical prompts describe subjects commonly found in traditional historical artworks, whereas contemporary prompts contain modern concepts that are largely absent from the historical artworks. Furthermore, each category is evaluated using both short prompts and detailed prompts to investigate the influence of prompt complexity on image generation quality. The generated results are assessed through qualitative visual inspection as well as quantitative evaluation using CLIP similarity and FID.

\subsection{Qualitative Results}

We first use short prompts and compare images generated by the original Stable Diffusion v1.5 model (base model) with those produced by LoRA-adapted checkpoints obtained from 1,000 to 5,000 training steps. Fig. \ref{fig:outputs} presents representative results generated from short prompts.

Compared with the base model, the fine-tuned checkpoints progressively exhibit visual characteristics that more closely resemble historical artworks. In particular, the generated images show changes in color composition, texture representation, brushwork appearance, and overall scene aesthetics. These stylistic characteristics become increasingly pronounced as the training progresses.

To evaluate whether the adapted model can generalize beyond the training distribution, we additionally generate images using contemporary concepts such as \textit{Tokyo Tower}. Since such concepts do not appear in the training dataset, successful generation requires the model to combine newly learned artistic characteristics with semantic knowledge retained from the original Stable Diffusion model. As illustrated in Figure~\ref{fig:outputs}, the adapted model is able to preserve the semantic identity of the target object while rendering it in a visual style that more closely resembles the historical artworks present in the training dataset.

\begin{figure*}[h]
\centering
\includegraphics[width=\textwidth]{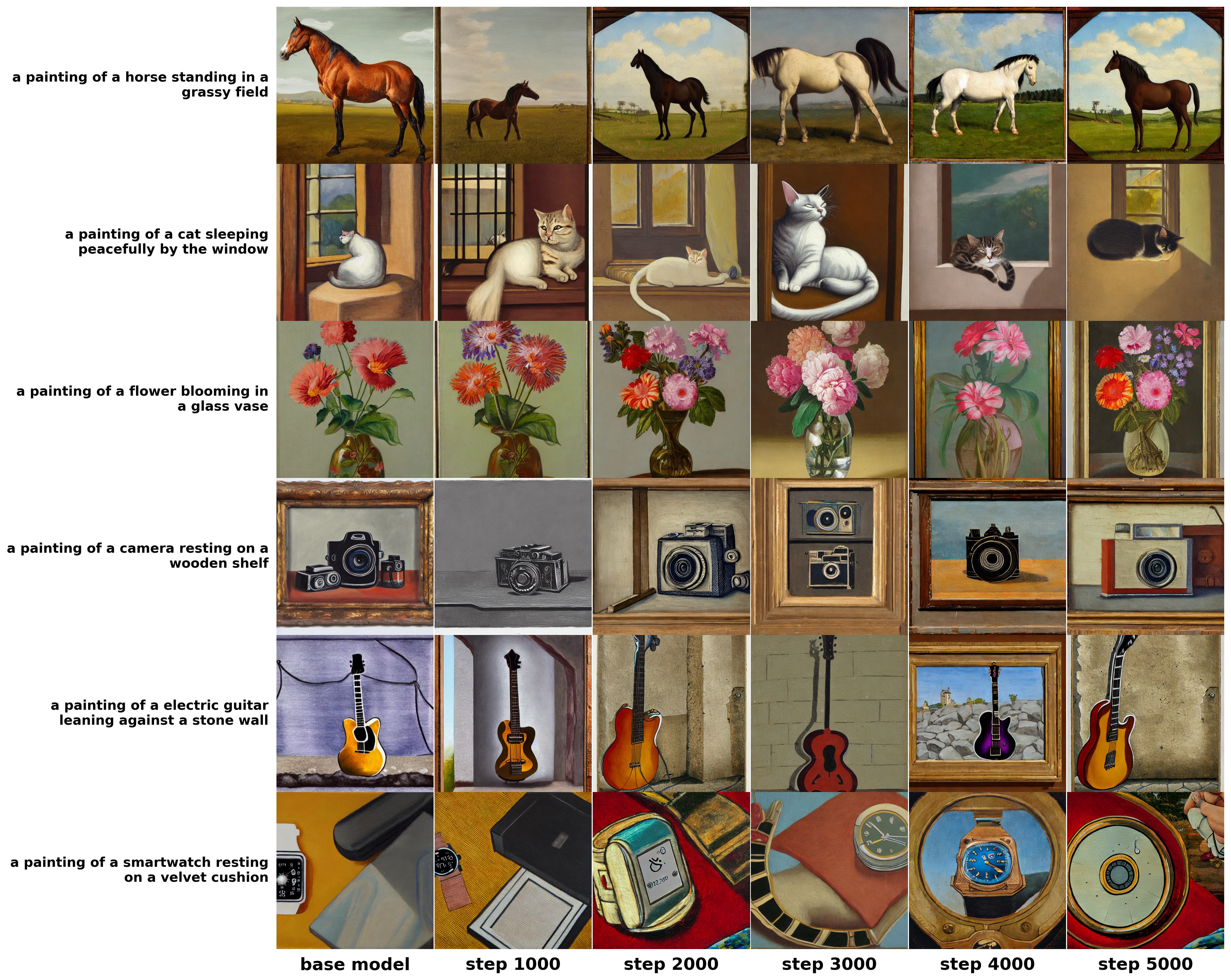}
\caption{Qualitative comparison using detailed long prompts. The adapted model preserves prompt semantics while exhibiting stronger historical-artistic visual characteristics than the base model.}
\label{fig:long outputs}
\end{figure*}

We further evaluate the model using detailed long prompts describing both historical and contemporary scenes. Representative examples are shown in Figure~\ref{fig:long outputs}. Similar trends can be observed across the generated images. The adapted model maintains semantic consistency with the input prompts while producing images whose visual appearance more closely resembles the artwork distribution represented by the training dataset. Notably, the model remains capable of generating modern objects such as cameras, electric guitars, and smart watches despite these concepts not being explicitly present in the training dataset.

\subsection{Quantitative Results}

To complement the qualitative analysis, we evaluate the generated images using CLIP similarity \cite{radford2021learning} and FID \cite{heusel2017gans}. For FID evaluation, the DEArt dataset serves as the reference image distribution. Results are reported separately for historical and contemporary prompts under both short and detailed prompt settings.

\subsubsection{CLIP Similarity}

\begin{table}[htbp]
\centering
\caption{CLIP similarity scores of the base model and the LoRA-adapted model after 5,000 training steps. Higher values indicate stronger text-image alignment.}
\label{tab:clip}
\small
\renewcommand{\arraystretch}{1.0}
\begin{tabular}{lcc}
\toprule
CLIP Similarity $\uparrow$ & Base Model & LoRA (5000 Steps) \\
\midrule
Short Historical Prompts & 31.407 & 30.311 \\
Short Contemporary Prompts & 31.080 & 29.654 \\
Long Historical Prompts & 34.610 & 34.430 \\
Long Contemporary Prompts & 35.820 & 34.740 \\
\bottomrule
\end{tabular}
\end{table}

Table~\ref{tab:clip} reports the average CLIP similarity scores. The results indicate that the fine-tuned model achieves slightly lower CLIP similarity scores across all prompt categories. This observation suggests a modest reduction in prompt-image alignment after adaptation. A possible explanation is that the LoRA fine-tuning process encourages the generated images to conform more closely to the stylistic characteristics of the DEArt dataset, introducing a trade-off between stylistic specialization and semantic alignment as measured by CLIP.\enlargethispage{16pt}

A second observation is that the detailed long prompts consistently produce higher CLIP similarity scores than short prompts for both the base and adapted models. This trend is expected because longer prompts provide richer semantic information, allowing both the diffusion model and the CLIP encoder to establish stronger text-image correspondences.

\subsubsection{FID}

\begin{table}[htbp]
\centering
\caption{FID scores of the base model and the LoRA-adapted model after 5,000 training steps. Lower values indicate greater similarity to the DEArt reference distribution.}
\label{tab:fid}
\small
\renewcommand{\arraystretch}{1.0}
\begin{tabular}{lcc}
\toprule
FID $\downarrow$ & Base Model & LoRA (5000 Steps) \\
\midrule
Short Historical Prompts & 230.11 & 207.57 \\
Short Contemporary Prompts & 252.08 & 229.18 \\
Long Historical Prompts & 230.91 & 221.45 \\
Long Contemporary Prompts & 270.17 & 256.66 \\
\bottomrule
\end{tabular}
\end{table}

Table~\ref{tab:fid} presents the FID results. Considering that the generated images are evaluated against a large and highly diverse artwork collection spanning multiple centuries and artistic styles, the absolute FID values are relatively high. Consequently, we focus primarily on the relative improvement between the base and adapted models.

Across all prompt categories, the LoRA-adapted model achieves lower FID scores than the base model. Since the DEArt dataset is used as the reference distribution, these improvements indicate that the generated images become more similar to the statistical distribution of historical artworks represented in DEArt. This observation is consistent with the qualitative results shown previously.

Historical prompts consistently achieve lower FID scores than contemporary prompts. This behavior is expected because many historical subjects share stronger semantic and visual similarities with the artworks contained in DEArt. In contrast, modern concepts such as electronic devices or contemporary landmarks are underrepresented or absent from the training dataset, making it more challenging for the adapted model to generate images that remain close to the DEArt distribution.

Interestingly, detailed prompts generally produce higher FID scores than short prompts. One possible explanation is that longer prompts encourage the generation of more complex scenes containing additional objects and interactions, thereby increasing the diversity of the generated image distribution. As a result, the generated samples may deviate further from the reference artwork distribution even when stylistic characteristics are preserved.

Overall, the quantitative results suggest that LoRA adaptation successfully shifts the generated image distribution toward the historical artwork domain represented by DEArt while maintaining relatively strong prompt-image alignment. The qualitative and quantitative analyses together indicate that the proposed framework can effectively incorporate historical-art visual characteristics into modern diffusion-based image generation models.

\section{Conclusion}

In this work, we presented \textit{The RenAIssance}, a framework for adapting text-to-image diffusion models through automatic caption generation and parameter-efficient fine-tuning. By combining BLIP-based caption synthesis with LoRA-based adaptation of Stable Diffusion, we transformed the DEArt collection into a large-scale image--text training dataset and investigated how effectively a modern diffusion model can generate historical-art visual media from historical artwork collections.

Through both qualitative and quantitative evaluations, we demonstrated that the adapted model successfully shifts the generated image distribution toward the visual characteristics represented in DEArt while maintaining the ability to synthesize diverse content from textual prompts. In particular, the model is capable of generating both historically relevant subjects and contemporary concepts by short and long prompts while preserving stylistic properties associated with historical paintings. The results further suggest that LoRA-based adaptation provides an efficient and practical approach for specializing large generative models to historical-art domains without requiring full model retraining.

More broadly, our findings indicate the potential of combining vision--language models and diffusion models for cultural and artistic visual media exploration. The proposed pipeline offers a scalable strategy for leveraging historical artwork collections that lack image--text annotations and demonstrates how automatically generated captions can facilitate domain-specific adaptation of generative models for the generation of historical-art visual media.

Several directions remain open for future investigation. First, the impact of automatically generated caption quality on stylistic learning and generation quality should be systematically evaluated. Comparing different caption-quality levels and captioning strategies could therefore provide insight into the sensitivity of the proposed pipeline to automatically generated annotations.

Second, the composition of the training dataset warrants further investigation. DEArt spans artworks from the twelfth to the eighteenth centuries and encompasses diverse artistic periods, subjects, styles, and modes of representation. While this diversity provides a broad foundation for learning historical aesthetics, it may also introduce substantial stylistic variation. Future work could therefore examine how dataset composition, temporal coverage, and stylistic diversity affect the resulting generations, including controlled experiments using subsets corresponding to particular periods, artists, or artistic characteristics.

Third, the choice of captioning and generative models represents another important research direction. Future experiments could compare alternative vision-language captioning models and more recent diffusion architectures to determine whether improvements in semantic understanding, image synthesis, or conditioning mechanisms lead to better preservation and controllability of historical artistic characteristics.

Last but not least, a systematic human-centered evaluation would provide a complementary perspective to purely computational or qualitative assessment. It is deserved to involve human participants and, in particular, European experts in vision for art and culture if available.

\begin{acks}
This work is co-funded by the European Union’s Horizon Europe research and innovation programme under Marie Skłodowska-Curie Actions (MSCA) with grant agreement No 101081674.

We thank the anonymous reviewers for their constructive comments and helpful suggestions.
\end{acks}



\end{document}